\documentclass{article} 
\usepackage{iclr2026_conference,times}

\usepackage{amsmath,amsfonts,bm}

\def\eqref#1{equation~\ref{#1}}

\def\1{\bm{1}}

\DeclareMathAlphabet{\mathsfit}{\encodingdefault}{\sfdefault}{m}{sl}
\SetMathAlphabet{\mathsfit}{bold}{\encodingdefault}{\sfdefault}{bx}{n}

\usepackage{amsmath,amssymb,amsthm}
\usepackage{algorithm}
\usepackage{algorithmic}
\usepackage{graphicx}
\usepackage{tikz}
\usepackage{standalone}
\usetikzlibrary{arrows.meta,decorations.pathreplacing,positioning}

\usepackage{hyperref}
\usepackage{url}

\title{DART: Decoded Attention over Recurrent States for Efficient Long-Context Sequence Modeling}

\author{%
  Yixiao Qian$^{1}$ \quad
  Song Chen$^{2,*}$ \quad
  Pengkai Wang$^{3}$ \quad
  Jiaxu Liu$^{4}$ \quad
  Shengze Cai$^{1,*}$ \quad
  Chao Xu$^{1,*}$ \\
  $^{1}$College of Control Science and Engineering, Zhejiang University\\
  $^{2}$Department of Mathematics, National University of Singapore\\
  $^{3}$Hong Kong Polytechnic University\\
  $^{4}$School of Science, Huzhou Normal University\\
  \texttt{yixiaoqian@zju.edu.cn, song.chen@nus.edu.sg, pengkai.wang@connect.polyu.hk}\\
  \texttt{jiaxuliu@zju.edu.cn, shengze\_cai@zju.edu.cn, cxu@zju.edu.cn}\\
  $^{*}$Corresponding authors.
}

\newif\ificlrpreprint
\iclrpreprinttrue
\ificlrpreprint
    \iclrfinalcopy
\fi
\begin{document}

\maketitle
\ificlrpreprint
    \lhead{Preprint}
\fi

\begin{abstract}
Modern language models are built primarily from Transformers, recurrent
models, and their hybrid architectures. Transformers rely on token-level attention
memories, while recurrent models such as state space models (SSMs) and linear
attention maintain compact recurrent states. These architectures are typically
instantiated separately or interleaved at the layer level, leaving open whether
a shared memory representation can support both recurrent compression and
attention-style retrieval.
We study this question through the state space duality (SSD) view of Mamba-2,
where the SSM state can be interpreted as a compressed associative key--value (KV) cache.
We observe that Mamba-2 decodes token-conditioned values from this
state but does not decode token-conditioned keys.
Based on this observation, we propose DART
(Decoded Attention over Recurrent sTates), which retains the chunk state
contributions produced by the Mamba-2 chunked scan as chunk state memories,
decodes token-conditioned keys and values from these memories, and performs
state-memory attention (SMA) over the resulting KV pairs. The retrieved
output is then combined with the native Mamba-2 output through a gated residual
connection.
DART supports practical training by reusing the Mamba-2 chunked scan and
implementing SMA as a FlashAttention-style computation.
Our analysis and experiments show that DART substantially reduces the
length-dependent inference cache compared with a matched attention baseline
(e.g., $75\%$ savings when the chunk size is $S=256$ and the state
size is $N=128$). Compared with Mamba-2, DART substantially improves
associative recall and retrieval while preserving general language-modeling
quality.
\end{abstract}

\section{Introduction}

Modern large language models (LLMs) are built mainly from Transformers,
recurrent models, and their hybrid architectures. Transformers are the dominant
backbone of LLMs and can directly attend to all previous
tokens~\citep{vaswani2017attention,brown2020language,touvron2023llama}.
However, full attention has quadratic training compute in sequence length, and
autoregressive inference requires a token-level KV cache that grows linearly
with context length. Recurrent models provide a representative alternative for
addressing this efficiency bottleneck, ranging from classical RNNs and LSTMs
to modern linear-time language models such as RWKV, RetNet, Mamba, and
DeltaNet~\citep{elman1990finding,hochreiter1997long,peng2023rwkv,sun2023retentive,gu2023mamba,dao2024transformers,yang2024parallelizing}.
These models process sequences through recurrent state updates and maintain a
compact recurrent state during inference.
While these advantages make recurrent models attractive for long sequences,
compressing long contexts into compact states creates a memory bottleneck.
Prior studies on associative recall, copying, and position-structured
retrieval show that recurrent models can lose long-range recall capacity
relative to attention~\citep{arora2024zoology,jelassi2024repeat,waleffe2024empirical}.

These complementary tradeoffs have motivated hybrid architectures that combine
attention and recurrent models. By mixing efficient recurrent layers with
attention layers, hybrid models can improve the trade-off between compute
efficiency and recall. Recent systems have achieved strong language-modeling
and retrieval performance~\citep{lieber2024jamba,glorioso2024zamba,ren2025samba,lahoti2026mamba}.
Most hybrid designs, however, keep token-level attention memories and compact
recurrent states as separate memory forms and combine them by interleaving or
stacking layers. This leaves open a more basic question: can the same memory
representation support both recurrent compression and attention-style
retrieval?

Mamba-2 provides a natural starting point through state space duality
(SSD)~\citep{dao2024transformers}. SSD shows that a selective state space model (SSM) layer can be
written as structured causal attention, connecting the SSM coefficients
$(A,B,C,X)$ with the attention roles of decay, keys, queries, and values. In this view,
the Mamba-2 state is a compressed KV cache. The recurrent update writes
token-dependent information into this cache, while the readout $C_tH_t$
extracts the value needed by the current token. SSD also leads to the chunked
scan algorithm of Mamba-2, which makes the recurrent computation efficient on
modern GPUs.

The key observation of this paper is that Mamba-2 reads values
from the compressed KV cache but does not extract keys from it. If keys can also
be decoded from this compressed state, then attention need not operate over all
token-level keys and values. It can instead retrieve from a much smaller set of
compressed memories produced by the recurrent scan. This
suggests a state-memory attention (SMA) mechanism between full token-level attention
and purely recurrent models: recurrent models learn to compress input tokens
into compact states, and attention retrieves from the
compressed memories according to the current query. 

Based on this observation,
we propose DART (Decoded Attention over Recurrent sTates), an architecture that
augments Mamba-2 with an SMA branch for attention-style retrieval over
compressed recurrent states. DART keeps the Mamba-2 chunked scan, treats each
chunk state contribution as a chunk state memory, extracts token-conditioned
keys and values from these memories, and adds the retrieved state readout as a
gated residual correction to the Mamba-2 output.
Our contributions can be summarized as follows:
\begin{itemize}
    \item We introduce DART, an architecture that connects recurrent compression
    with attention-style retrieval through SMA. The Mamba-2 scan produces chunk
    state contributions that serve as chunk state memories, and SMA retrieves
    from them with token-conditioned keys and values.
    \item We show that DART is practical to train by reusing the
    Mamba-2 chunked scan and implementing SMA with a
    FlashAttention-style computation~\citep{dao2022flashattention,dao2024flashattention}.
    \item We evaluate DART on MQAR, NIAH, real-world retrieval tasks,
    Pile pretraining, and zero-shot downstream evaluations. Compared with
    Mamba-2, DART substantially improves associative recall and long-context
    retrieval while preserving language-modeling quality. Compared with matched 
    attention, DART substantially reduces the length-dependent inference cache
    without relying on sparse or sliding-window mechanism.
\end{itemize}

\section{Background}

\paragraph{State Space Models.}
Inspired by continuous dynamical systems, state space models (SSMs) map a
sequence $\boldsymbol{x}\in\mathbb{R}^{L}$ to
$\boldsymbol{y}\in\mathbb{R}^{L}$ through an implicit latent state
$\boldsymbol{h}$:
\begin{equation}
    \label{eq:ssm_ode}
    \dot{\boldsymbol{h}}(t) = \boldsymbol{A} \boldsymbol{h}(t) + \boldsymbol{B} x(t),
    \qquad
    y(t) = \boldsymbol{C} \boldsymbol{h}(t).
\end{equation}
Modern SSMs can be broadly grouped into non-selective and
selective variants. 
Non-selective SSMs such as S4, S4D, and S5 use
input-independent dynamics and exploit their structure for efficient convolution
or scan algorithms~\citep{gu2021efficiently,gu2022parameterization,smith2022simplified}.
Selective SSMs such as Mamba, Mamba-2, and Mamba-3 make the dynamics
input-dependent, allowing the model to choose what to write, retain, and read
from the state at each token~\citep{gu2023mamba,dao2024transformers,lahoti2026mamba}.

\paragraph{State Space Duality.}
Mamba-2 introduces state space duality (SSD), which connects selective SSMs
and attention through an equivalent structured causal attention form. For
$X,Y\in\mathbb{R}^{L\times P}$ and scalar transition $A_t=a_tI_N$, Mamba-2 
can be written as
\begin{equation}
    H_t
    =
    A_tH_{t-1}
    +
    B_t^\top X_t,
    \qquad
    Y_t=C_tH_t ,
\end{equation}
where $H_t\in\mathbb{R}^{N\times P}$, $B_t \in\mathbb{R}^{1\times N}$, and $C_t \in\mathbb{R}^{1\times N}$. 
The update $B_t^\top X_t$ compresses the current input $X_t$ into the state $H_t$,
while the readout $C_tH_t$ decodes the value needed by the current token from
this state.
We write
\begin{equation}
    A_{u:v}
    =
    A_vA_{v-1}\cdots A_u
    =
    \prod_{k=u}^{v}A_k,
    \qquad u\leq v,
\end{equation}
and set $A_{u:v}=I_N$ when $u>v$.
Let $B,C\in\mathbb{R}^{L\times N}$ stack the token-wise vectors $B_t$ and
$C_t$ row-wise. Without the SSD decay, the product $CB^\top X$ has the same algebraic
form as linear attention, namely $QK^\top V$ with $Q=C$, $K=B$, and $V=X$.
SSD adds a structured causal decay mask to this kernel and writes the Mamba-2
layer as structured causal attention:
\begin{equation}
    Y
    =
    \left(
    L_{\mathrm{SSD}}\circ CB^\top
    \right)X,
    \qquad
    Q=C,\quad K=B,\quad V=X .
\end{equation}
Here $(L_{\mathrm{SSD}})_{ij} = \prod_{k=j+1}^{i}a_k$ for $i\geq j$,
and $(L_{\mathrm{SSD}})_{ij}=0$ for $i<j$. 
SSD shows that Mamba-2 can be viewed as an iterative form of structured
attention. In this view, $H_t$ is a compressed KV cache, and $C_tH_t$ extracts
the value needed by the current token from this cache.

\paragraph{Memory Issues of Recurrent Models.}
Recurrent models reduce memory cost by compressing history into compact states,
but this compression creates a bottleneck for precise long-range recall. In
selective SSMs, for example, selectivity can be viewed as a learned compression
policy: during training, the model learns which tokens should be preserved in
the state and which tokens can be forgotten. This compression is powerful but
learned from training data. When a past token is processed, the model cannot
know which future query will need it. A straightforward way to store more
history is to use a larger state, which increases memory capacity but does not
fundamentally address the lack of attention-like search over past tokens.
Several studies have identified this limitation on associative recall, copying,
and position-structured retrieval, where recurrent models can have weaker
long-range recall than attention~\citep{arora2024zoology,jelassi2024repeat,waleffe2024empirical}.
Recent work has also developed more explicit memory-editing mechanisms:
delta-rule recurrent models improve compact states by subtracting stale reads,
applying targeted updates, combining such updates with adaptive gating, or
decoupling erase and write
controls~\citep{yang2024parallelizing,yang2025gated,hatamizadeh2026gated}.

\paragraph{Bottlenecks of Long-Context Attention.}
Modern LLMs are built primarily on the Transformer architecture, but extending
Transformers to long contexts remains challenging. 
Many efforts have therefore focused on reducing the training and inference
costs of long-context attention.
FlashAttention exploits the GPU
memory hierarchy to remove the quadratic activation-memory cost of traditional
attention~\citep{dao2022flashattention,dao2024flashattention}, but very
long-context inference is still limited by token-level KV-cache memory and
bandwidth. Existing systems reduce KV-cache cost through
sliding-window or block-sparse patterns~\citep{beltagy2020longformer,zaheer2020big,xiao2024efficient},
dynamic cache selection during decoding~\citep{zhang2023h2o,li2024snapkv},
latent KV representations such as multi-head latent attention (MLA)~\citep{liu2024deepseek},
or KV-cache quantization~\citep{liu2024kivi}. 
While effective, these approaches mainly optimize, approximate, or redesign the token-level KV cache itself, rather than exploiting the compact recurrent state already maintained by the model.

\begin{figure}[htbp]
    \centering
    \begin{tikzpicture}[
    outer/.style={draw, fill=black!4, rounded corners=15pt, line width=1.15pt,
        minimum width=3.0cm, minimum height=3.05cm},
    block/.style={draw, rounded corners=5pt, line width=1.05pt, align=center,
        minimum width=2.25cm, minimum height=0.62cm,
        inner sep=3pt, font=\scriptsize},
    attn/.style={block, fill=red!9},
    ssm/.style={block, fill=yellow!13},
    sma/.style={block, fill=blue!11},
    residual/.style={line width=1.05pt, rounded corners=5pt},
    oplus/.style={
        draw=black, line width=0.8pt, circle, minimum size=8pt,
        inner sep=0pt, outer sep=0pt,
        path picture={
            \draw (path picture bounding box.center) -- ++(0.3cm,0)
                (path picture bounding box.center) -- ++(-0.3cm,0)
                (path picture bounding box.center) -- ++(0,0.3cm)
                (path picture bounding box.center) -- ++(0,-0.3cm);
        },
    },
    matcell/.style={draw=black!55, fill=yellow!12, line width=0.45pt,
        minimum width=0.38cm,
        minimum height=0.30cm, inner sep=0pt},
    vcell/.style={draw=black!55, fill=blue!10, line width=0.45pt,
        minimum width=0.38cm,
        minimum height=0.24cm, inner sep=0pt},
    kcell/.style={draw=black!55, fill=red!8, line width=0.45pt,
        minimum width=0.24cm,
        minimum height=0.30cm, inner sep=0pt},
    title/.style={font=\small\bfseries, align=center},
    note/.style={font=\scriptsize, align=center},
    arrow/.style={-{Latex[length=2.2mm, width=1.5mm]}, line width=1.05pt},
    looparrow/.style={-{Latex[length=2.2mm, width=1.5mm]}, line width=1.05pt},
    brace/.style={decorate, decoration={brace, amplitude=4pt}, line width=0.9pt}
]
    \node[title] at (0, 2.65) {Interleaved hybrid};
    \node[outer, minimum width=3.35cm, minimum height=3.65cm]
        (left_outer) at (0, 0) {};
    \node[ssm] (left_ssm) at (0, -0.92) {SSM block};
    \node[oplus] (left_ssm_add) at (0, -0.20) {};
    \node[attn] (left_swa) at (0, 0.77) {SWA block};
    \node[oplus] (left_swa_add) at (0, 1.50) {};
    \draw[arrow] (0, -2.25) -- (left_ssm.south);
    \draw[line width=1.05pt] (left_ssm.north) -- (left_ssm_add.south);
    \draw[arrow] (left_ssm_add.north) -- (left_swa.south);
    \draw[line width=1.05pt] (left_swa.north) -- (left_swa_add.south);
    \draw[arrow] (left_swa_add.north) -- (0, 2.25);
    \draw[residual] (0, -1.55)
        -- (-1.34, -1.55)
        -- (-1.34, -0.20)
        -- (left_ssm_add.center);
    \draw[residual] (0, 0.02)
        -- (-1.34, 0.02)
        -- (-1.34, 1.50)
        -- (left_swa_add.center);

    \node[title] at (4.1, 2.65) {DART memory-level hybrid};
    \node[outer, minimum width=3.35cm, minimum height=3.65cm]
        (right_outer) at (4.1, 0) {};
    \coordinate (right_split) at (4.1, -1.02);
    \node[ssm, minimum width=1.05cm, minimum height=0.56cm]
        (right_ssm) at (3.4, -0.20) {SSM};
    \node[sma, minimum width=1.05cm, minimum height=0.56cm]
        (right_sma) at (4.8, -0.20) {SMA};
    \node[oplus] (right_memory_add) at (4.1, 0.68) {};
    \node[oplus] (right_residual_add) at (4.1, 1.45) {};
    \draw[line width=1.05pt] (4.1, -2.25) -- (right_split);
    \draw[arrow] (right_split) -- (3.4, -1.02) -- (right_ssm.south);
    \draw[arrow] (right_split) -- (4.8, -1.02) -- (right_sma.south);
    \draw[arrow] (right_ssm.east) -- (right_sma.west);
    \draw[arrow] (right_ssm.north)
        -- (3.4, 0.68)
        -- (right_memory_add.west);
    \draw[arrow] (right_sma.north)
        -- (4.8, 0.68)
        -- (right_memory_add.east);
    \draw[line width=1.05pt]
        (right_memory_add.north) -- (right_residual_add.south);
    \draw[arrow] (right_residual_add.north) -- (4.1, 2.25);
    \draw[residual] (4.1, -1.50)
        -- (2.76, -1.50)
        -- (2.76, 1.45)
        -- (right_residual_add.center);

    \node[note, text width=3.15cm] at (0, -2.65)
        {Attention and recurrent modules are inserted as separate layers.};
    \node[note, text width=3.25cm] at (4.15, -2.65)
        {Attention-style retrieval is performed over chunk state memories.};

    \node[title] at (9.50, 2.65)
        {Chunk state memory $\Delta H_{[c]}$};
    \node[outer, minimum width=6.50cm, minimum height=3.65cm]
        (matrix_outer) at (9.50, 0) {};
    \foreach \c in {0,...,5} {
        \node[vcell] at (8.05 + 0.38*\c, 1.20) {};
    }
    \node[font=\scriptsize] at (9.0, 1.57)
        {$V_{t,c}\in\mathbb{R}^{P}$};

    \foreach \r in {0,...,4} {
        \foreach \c in {0,...,5} {
            \node[matcell] at (8.05 + 0.38*\c, 0.57 - 0.30*\r) {};
        }
    }
    \draw[decorate, decoration={brace, mirror, amplitude=4pt}, line width=0.9pt]
        (7.83, -1.11) --
        node[below=4pt, note] {value features $P$} (10.17, -1.11);
    \draw[decorate, decoration={brace, mirror, amplitude=4pt}, line width=0.9pt]
        (7.72, 0.72) -- (7.72, -0.78);
    \node[note, align=center] at (7.05, -0.03) {key\\features $N$};

    \draw[arrow] (9.0, 0.79) -- node[right, note] {$C_t$} (9.0, 1.05);

    \foreach \r in {0,...,4} {
        \node[kcell] at (11.08, 0.57 - 0.30*\r) {};
    }
    \node[font=\scriptsize] at (11.88, -0.03)
        {$K_{t,c}\in\mathbb{R}^{N}$};
    \draw[arrow] (10.18, -0.03) -- node[above, note] {$E_t$} (10.94, -0.03);

    \node[note, text width=4.5cm] at (9.50, -2.65)
        {The same chunk state memory decodes values through $C_t$ and
        token-conditioned keys through $E_t$.};
\end{tikzpicture}
    \caption{Comparison between interleaved hybrid architectures and DART.
    Standard interleaved hybrids place attention and recurrent modules in
    separate layers, whereas DART integrates attention-style retrieval within
    each recurrent block through SMA over chunk state memories. The matrix view
    shows that a chunk state contribution
    $\Delta H_{[c]}\in\mathbb{R}^{N\times P}$ supports both value readout
    $C_t\Delta H_{[c]}$ and token-conditioned key decoding $\Delta H_{[c]}E_t$.}
    \label{fig:hybrid_design_comparison}
\end{figure}
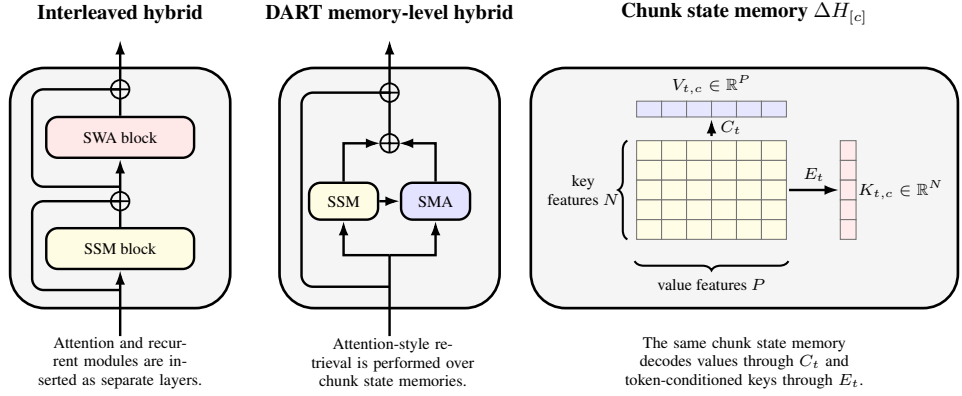

\paragraph{Hybrid Architectures.}
The complementary strengths of attention and recurrent models have motivated
hybrid sequence architectures. Existing interleaved systems usually place the
two components in separate layers, and their attention--recurrent ratio is
typically selected empirically, as in representative systems such as Jamba,
Zamba, and Samba~\citep{lieber2024jamba,glorioso2024zamba,ren2025samba}.
Recent chunk-level attention methods have also moved beyond token-level KV
memory. Attamba and RAT both perform attention over chunk-level
representations: Attamba uses SSMs to compress token chunks into key-value
states, while RAT applies recurrence within chunks and softmax attention across
chunks~\citep{akhauri2024attamba,wei2025rat}. These methods mainly focus on
using recurrent or SSM modules to produce compressed K/V for an attention layer.
Figure~\ref{fig:hybrid_design_comparison} illustrates the distinction between DART and interleaved hybrid architectures:
interleaved hybrids insert attention and recurrent modules as separate layers,
whereas DART performs attention-style retrieval inside the recurrent block by
reading its chunk state memories.

\section{Methodology}
\label{sec:delay_mamba}

DART augments the Mamba-2 chunked scan with state-memory attention (SMA) over
its chunk state contributions, which DART retains as chunk state memories. In
the SSD view, the recurrent state $H_t$ is a compressed KV cache: Mamba-2 reads
values through $C_tH_t$, while DART also decodes token-conditioned keys from
chunk state memories. We refer to the original Mamba-2 scan and readout as the
SSM branch, and to the retrieval module over chunk state memories as the SMA
branch. More details about Mamba-2 chunked scan are provided in Appendix~\ref{app:mamba2}.

\paragraph{Chunk State Memories.}
For a sequence of length $L$ and chunk size $S$, DART divides the sequence
into chunks. We write $[c]$ for the set of token indices in the $c$-th chunk,
let $c(t)=\lceil t/S\rceil$ denote the chunk containing token $t$, and let
$\Delta H_{[c]}$ denote its chunk state contribution:
\begin{equation}
    \Delta H_{[c]}
    =
    \sum_{s\in[c]}
    A_{s+1:e(c)}B_s^\top X_s
    \in\mathbb{R}^{N\times P}.
\end{equation}
Here $e(c)$ is the last index in $[c]$, and
$A_{s+1:e(c)}=\left(\prod_{k=s+1}^{e(c)}a_k\right) I_N$, with the empty product equal to
the identity.
In the recurrent scan, we refer to $\Delta H_{[c]}$ as a \emph{chunk state
contribution} because it is the contribution of chunk $c$ to the
chunk-boundary state. When retained for SMA, the same tensor serves as a
\emph{chunk state memory}.
The matrix $\Delta H_{[c]}$ has a key axis $N$ and a value axis
$P$. 
Reading it from the left with $C_t\in\mathbb{R}^{1\times N}$ produces a value,
while contracting its value axis with $E_t\in\mathbb{R}^{P\times 1}$ produces
a key:
\begin{equation}
    C_t\Delta H_{[c]}
    \in\mathbb{R}^{1\times P}
    \quad\text{(value-side readout)},
    \qquad
    \Delta H_{[c]}E_t
    \in\mathbb{R}^{N\times 1}
    \quad\text{(key-side readout)}.
\end{equation}

\paragraph{State-Memory Attention.}
Let $U_t\in\mathbb{R}^{1\times d_{\mathrm{model}}}$ denote the input
representation to the DART block.
For a single SSM head and token $t$, DART forms three token-conditioned
vectors from the block input $U_t$:
\begin{equation}
    Q_t
    =
    U_tW_Q
    \in\mathbb{R}^{1\times N},
    \qquad
    C_t
    =
    U_tW_C
    \in\mathbb{R}^{1\times N},
    \qquad
    E_t
    =
    \left(U_tW_E\right)^\top
    \in\mathbb{R}^{P\times 1},
\end{equation}
where $W_Q,W_C\in\mathbb{R}^{d_{\mathrm{model}}\times N}$ and
$W_E\in\mathbb{R}^{d_{\mathrm{model}}\times P}$.
Here $Q_t$ is a query vector in the SSD key-side space and is used for chunk routing.
The vector $C_t$ is the native Mamba-2 read vector shared with the SMA branch,
and $C_t\Delta H_{[c]}$ extracts the value needed by token $t$ from the chunk
state memory.
Empirically, this sharing is important for optimization, as an independent SMA
read vector did not learn useful recall behavior in our experiments.
The vector $E_t$ is a value-side verifier, and
$\Delta H_{[c]}E_t$ turns each chunk state memory into a
token-conditioned key.

For each historical chunk $c$, DART computes:
\begin{equation}
    V_{t,c}
    =
    C_t\Delta H_{[c]}
    \in\mathbb{R}^{1\times P},
    \qquad
    K_{t,c}
    =
    \Delta H_{[c]}E_t
    \in\mathbb{R}^{N\times 1}.
\end{equation}
And the chunk logit is:
\begin{equation}
    \ell_{t,c}
    =
    \frac{Q_tK_{t,c}}{\sqrt{N}}
    \in\mathbb{R}.
\end{equation}

\paragraph{Gated Residual Integration.}
The SMA logits are normalized over historical chunks:
\begin{equation}
    g_{t,c}
    =
    \operatorname{softmax}_{c<c(t)}(\ell_{t,c}),
    \qquad
    R_t
    =
    \sum_{c<c(t)}g_{t,c}V_{t,c}
    \in\mathbb{R}^{1\times P}.
\end{equation}
For tokens in the first chunk, the historical memory set is empty; in this
case we define the SMA readout as $R_t=0$.
In practice, the score branch uses RMS-normalized $Q_t$, $E_t$, and chunk
states for stable routing, with details provided in
Appendix~\ref{app:method_details}. DART then adds the SMA readout
as a scalar-gated residual correction to the SSM branch readout:
\begin{equation}
    G_t = \operatorname{SiLU}(U_tW_G)\in\mathbb{R},
    \qquad
    Y_t = C_tH_t + G_t R_t,
\end{equation}
where $W_G\in\mathbb{R}^{d_{\mathrm{model}}\times 1}$.
We initialize $W_G$ to zero, so the SMA residual starts from $G_tR_t=0$ and
is learned as a residual correction during training.

\begin{figure}[htbp]
    \centering
    \includegraphics[width=0.7\linewidth]{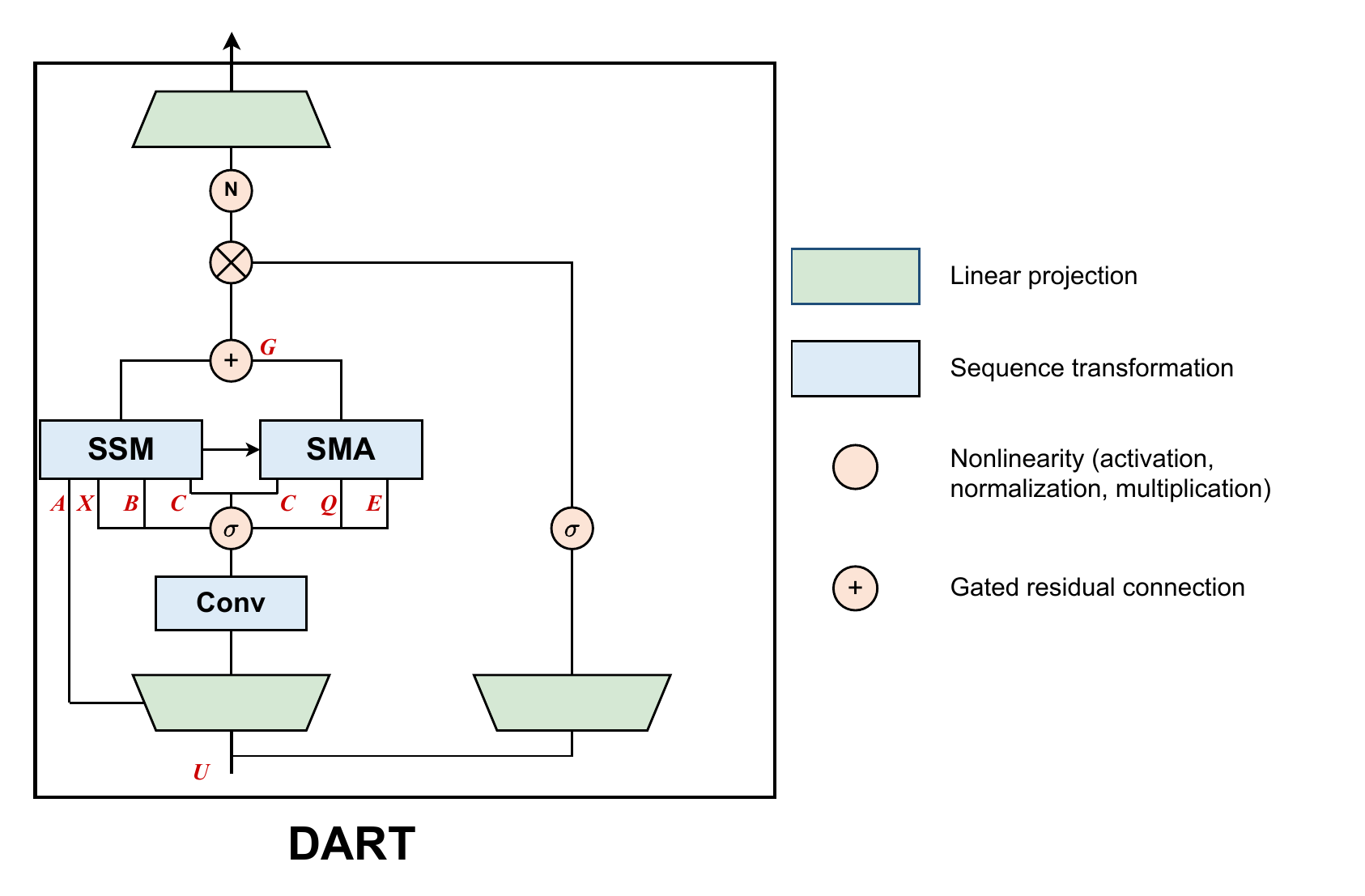}
    \caption{Architecture of DART. The SSM branch provides chunk state
    memories to the SMA branch, whose retrieved readout is combined with
    the SSM output through a gated residual connection. In practice, the $B$, $C$, $Q$, and
    $E$ projections use a single projection group ($G=1$) shared across all heads, analogous
    to multi-value attention (MVA) in Mamba-2.}
    \label{fig:delay_mamba_architecture}
\end{figure}

The architecture of DART is shown in
Figure~\ref{fig:delay_mamba_architecture}. The SSM branch follows the Mamba-2
block and passes chunk state memories to the SMA
branch. The SMA branch uses token-conditioned $Q_t$ and $E_t$ for retrieval, and
its readout is added to the SSM output through a gated residual connection.

\section{Efficient Training and Complexity Analysis}
\label{sec:efficient_kernels}

DART can be trained efficiently because it reuses the Mamba-2 chunked
scan to produce chunk state memories, and implements SMA with
FlashAttention-style streaming and recomputation. The complexity analysis
shows that DART can substantially reduce inference cache size. With a
common setting of chunk size $S=256$ and state size $N=128$, DART saves
75\% cache memory compared with token-level attention.

\begin{figure}[H]
    \centering
    \definecolor{scanorange}{HTML}{E67E22}
\definecolor{stateyellow}{HTML}{E5B800}
\definecolor{stateblue}{HTML}{5576C2}
\definecolor{writegreen}{HTML}{6AA84F}
\definecolor{stategray}{HTML}{B8B8B8}
\definecolor{smaaccent}{HTML}{3D8B85}

\begin{tikzpicture}[
    tokenrow/.style={
        draw=black,
        line width=0.75pt,
        minimum width=1.92cm,
        minimum height=0.48cm,
        inner sep=0pt,
        fill=white
    },
    boundary/.style={
        draw=black,
        dashed,
        line width=0.8pt,
        minimum width=0.38cm,
        minimum height=0.82cm,
        inner sep=0pt,
        fill=stategray!22
    },
    memorybox/.style={
        draw=black,
        dashed,
        line width=0.8pt,
        minimum width=1.45cm,
        minimum height=0.48cm,
        inner sep=0pt,
        font=\scriptsize,
        text=writegreen,
        fill=stategray!22
    },
    decodedkey/.style={
        draw=stateblue,
        line width=0.7pt,
        minimum width=0.72cm,
        minimum height=0.44cm,
        inner sep=0pt,
        font=\tiny,
        fill=stateblue!13
    },
    decodedvalue/.style={
        draw=scanorange,
        line width=0.7pt,
        minimum width=0.72cm,
        minimum height=0.44cm,
        inner sep=0pt,
        font=\tiny,
        fill=scanorange!11
    },
    smabox/.style={
        draw=smaaccent,
        rounded corners=1.5pt,
        line width=0.8pt,
        minimum width=6.10cm,
        minimum height=0.58cm,
        align=center,
        font=\scriptsize,
        fill=smaaccent!8
    },
    intra/.style={-{Straight Barb}, draw=scanorange, line width=1.15pt},
    write/.style={-{Straight Barb}, draw=writegreen, line width=1.15pt},
    read/.style={-{Straight Barb}, draw=stateblue, line width=1.15pt},
    scan/.style={-{Straight Barb}, draw=stateyellow, line width=1.15pt},
    smaflow/.style={-{Straight Barb}, draw=smaaccent, line width=1.15pt},
    rowlabel/.style={font=\small\bfseries, anchor=east},
    rowsymbol/.style={font=\small, anchor=west, text=black!78},
    paneltitle/.style={font=\small\bfseries}
]
    \node[paneltitle] at (5.15,2.60) {(a) Mamba-2 chunked scan};

    \node[rowlabel] at (0.95,1.90) {Outputs};
    \node[rowsymbol] at (1.08,1.90) {$Y$};
    \node[rowlabel, text=black!42] at (0.85,0.88) {States};
    \node[rowsymbol, text=black!42] at (0.98,0.88) {$H_{[c]}$};
    \node[rowlabel] at (0.95,0) {Inputs};
    \node[rowsymbol] at (1.08,0) {$X$};

    \foreach \x/\name in {2.75/one,5.65/two,8.55/three} {
        \node[tokenrow] (x\name) at (\x,0) {};
        \node[tokenrow] (y\name) at (\x,1.90) {};
        \foreach \shift in {-0.48,0,0.48} {
            \draw[line width=0.65pt]
                ([xshift=\shift cm]x\name.south) --
                ([xshift=\shift cm]x\name.north);
            \draw[line width=0.65pt]
                ([xshift=\shift cm]y\name.south) --
                ([xshift=\shift cm]y\name.north);
        }
    }

    \node[boundary] (hone) at (4.20,0.95) {};
    \node[boundary] (htwo) at (7.10,0.95) {};
    \node[boundary] (hthree) at (10.00,0.95) {};

    \node[font=\scriptsize, text=black!62, below=2pt of xone] {chunk $1$};
    \node[font=\scriptsize, text=black!62, below=2pt of xtwo] {chunk $2$};
    \node[font=\scriptsize, text=black!62, below=2pt of xthree] {chunk $3$};

    \draw[intra] (2.75,0.42) -- (2.75,1.48);
    \draw[intra] (5.65,0.42) -- (5.65,1.48);
    \draw[intra] (8.55,0.42) -- (8.55,1.48);

    \draw[write]
        (3.82,0.05) to[out=0,in=-90] (4.20,0.43);
    \draw[write]
        (6.72,0.05) to[out=0,in=-90] (7.10,0.43);
    \draw[write]
        (9.62,0.05) to[out=0,in=-90] (10.00,0.43);
    \node[font=\scriptsize, text=writegreen, anchor=north]
        at (4.17,-0.02) {$\Delta H_{[1]}$};
    \node[font=\scriptsize, text=writegreen, anchor=north]
        at (7.07,-0.02) {$\Delta H_{[2]}$};
    \node[font=\scriptsize, text=writegreen, anchor=north]
        at (9.97,-0.02) {$\Delta H_{[3]}$};

    \draw[read]
        (4.20,1.47) to[out=90,in=180] (4.58,1.90);
    \draw[read]
        (7.10,1.47) to[out=90,in=180] (7.48,1.90);

    \draw[scan]
        (4.52,0.49) .. controls (5.05,0.34) and (6.25,0.34) .. (6.78,0.49);
    \draw[scan]
        (7.42,0.49) .. controls (7.95,0.34) and (9.15,0.34) .. (9.68,0.49);

    \node[paneltitle] at (14.00,2.60) {(b) State-memory attention};

    \node[memorybox] (mone) at (11.70,0) {$\Delta H_{[1]}$};
    \node[memorybox] (mtwo) at (14.00,0) {$\Delta H_{[2]}$};
    \node[memorybox] (mthree) at (16.30,0) {$\Delta H_{[3]}$};

    \foreach \memory/\x/\idx in {mone/11.70/1,mtwo/14.00/2,mthree/16.30/3} {
        \node[decodedkey] (k\idx) at ({\x-0.43},0.84) {$K_{t,\idx}$};
        \node[decodedvalue] (v\idx) at ({\x+0.43},0.84) {$V_{t,\idx}$};

        \draw[read] ([xshift=-0.43cm]\memory.north) -- (k\idx.south);
        \draw[intra] ([xshift=0.43cm]\memory.north) -- (v\idx.south);
        \node[font=\tiny, text=stateblue, anchor=east]
            at ({\x-0.51},0.49) {$E_t$};
        \node[font=\tiny, text=scanorange, anchor=west]
            at ({\x+0.51},0.49) {$C_t$};
    }

    \node[smabox] (sma) at (14.00,1.65)
        {\textbf{State-Memory Attention}};
    \foreach \idx in {1,2,3} {
        \draw[read] (k\idx.north) -- (k\idx.north |- sma.south);
        \draw[intra] (v\idx.north) -- (v\idx.north |- sma.south);
    }

    \draw[smaflow] (sma.north) -- ++(0,0.28);

\end{tikzpicture}
    \caption{Construction and retrieval of chunk state memories in DART.
    \textbf{Left}: the
    Mamba-2 chunked scan computes local outputs and chunk state contributions
    while propagating compact boundary states. \textbf{Right}: for each token
    $t$, $E_t$ and $C_t$ decode a key $K_{t,c}$ and value $V_{t,c}$ from each
    historical chunk state memory. SMA attends over the decoded key--value
    pairs and produces the readout $R_t$.}
    \label{fig:delay_mamba_efficient_training}
\end{figure}
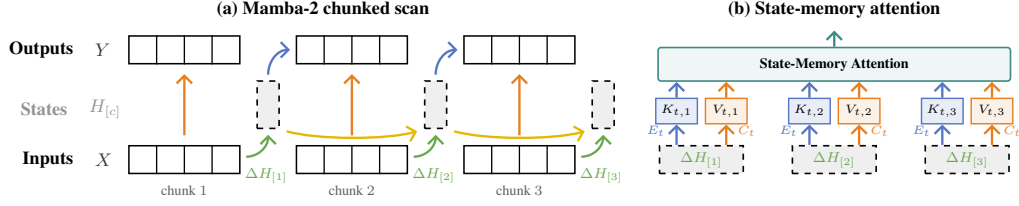

\paragraph{Efficient Training.}
Figure~\ref{fig:delay_mamba_efficient_training} illustrates the Mamba-2
chunked scan and SMA readout used by DART.
DART reuses the efficient Mamba-2 chunked SSD scan
~\citep{dao2024transformers}, which produces both the SSM branch output $C_tH_t$
and the chunk state contributions $\Delta H_{[c]}$.
The SMA branch is implemented with a custom FlashAttention-style
kernel~\citep{dao2022flashattention,dao2024flashattention}.
Following FlashAttention, the kernel fixes a query tile and streams over
historical chunks. For each historical chunk $c$, the
kernel constructs $K_{t,c}$ and $V_{t,c}$ from $\Delta H_{[c]}$ on chip,
updates the online softmax over chunks, and accumulates the SMA readout
$R_t$. Appendix~\ref{app:flash_attention} provides the background on online
softmax and recomputation.

\begin{algorithm}[htbp]
\caption{FlashAttention-Style State-Memory Attention}
\label{alg:state_memory_readout}
\begin{algorithmic}[1]
\REQUIRE Matrices $Q,C\in\mathbb{R}^{L\times N}$,
$E\in\mathbb{R}^{P\times L}$, chunk state memories
$\Delta H_{[1]},\ldots,\Delta H_{[M]}\in
\mathbb{R}^{N\times P}$ in HBM, chunk size $S$, token tile size $T$.
\ENSURE SMA readout $R\in\mathbb{R}^{L\times P}$ and logsumexp
$L_{\mathrm{lse}}\in\mathbb{R}^{L}$.
\STATE Divide the sequence into $M=\lceil L/S\rceil$ chunks; divide each
chunk into token tiles of size $T$.
\FOR{target chunk $i=1,\ldots,M$}
    \FOR{token tile $\tau$ inside chunk $i$}
        \STATE Load $Q_\tau,C_\tau\in\mathbb{R}^{T\times N}$ and
        $E_\tau\in\mathbb{R}^{P\times T}$ from HBM to on-chip SRAM.
        \STATE On chip, initialize
        $O_\tau^{(0)}=0\in\mathbb{R}^{T\times P}$,
        $d_\tau^{(0)}=0\in\mathbb{R}^{T}$, and
        $m_\tau^{(0)}=-\infty\in\mathbb{R}^{T}$.
        \FOR{historical chunk $j<i$}
            \STATE Load $\Delta H_{[j]}$ from HBM to on-chip SRAM.
            \STATE On chip, compute
            $K_\tau^{(j)}=(\Delta H_{[j]}E_\tau)^\top
            \in\mathbb{R}^{T\times N}$.
            \STATE On chip, compute
            $\ell_\tau^{(j)}
            =
            \operatorname{rowsum}(Q_\tau\odot K_\tau^{(j)})/\sqrt N
            \in\mathbb{R}^{T}$.
            \STATE On chip, compute
            $V_\tau^{(j)}=C_\tau\Delta H_{[j]}
            \in\mathbb{R}^{T\times P}$.
            \STATE On chip, compute
            $m_\tau^{(j)}=\max(m_\tau^{(j-1)},\ell_\tau^{(j)})$ and
            $\widetilde g_\tau^{(j)}=\exp(\ell_\tau^{(j)}-m_\tau^{(j)})$.
            \STATE On chip, update
            $d_\tau^{(j)}
            =
            e^{m_\tau^{(j-1)}-m_\tau^{(j)}}d_\tau^{(j-1)}
            +
            \widetilde g_\tau^{(j)}$.
            \STATE On chip, update
            $O_\tau^{(j)}
            =
            \operatorname{diag}(e^{m_\tau^{(j-1)}-m_\tau^{(j)}})
            O_\tau^{(j-1)}
            +
            \operatorname{diag}(\widetilde g_\tau^{(j)})V_\tau^{(j)}$.
        \ENDFOR
        \STATE On chip, compute
        $R_\tau=\operatorname{diag}(d_\tau^{(i-1)})^{-1}
        O_\tau^{(i-1)}$, keeping $R_\tau=0$ when no historical chunk exists.
        \STATE On chip, compute
        $(L_{\mathrm{lse}})_\tau
        =
        m_\tau^{(i-1)}+\log d_\tau^{(i-1)}$, keeping
        $(L_{\mathrm{lse}})_\tau=-\infty$ when no historical chunk exists.
        \STATE Write $R_\tau$ to HBM as the corresponding tile of $R$.
        \STATE Write $(L_{\mathrm{lse}})_\tau$ to HBM as the corresponding
        tile of $L_{\mathrm{lse}}$.
    \ENDFOR
\ENDFOR
\STATE Return the SMA readout $R$ and logsumexp $L_{\mathrm{lse}}$.
\end{algorithmic}
\end{algorithm}

Algorithm~\ref{alg:state_memory_readout} summarizes this forward kernel.
For clarity, the algorithm omits the RMS normalization applied to the SMA
score branch; the normalized formulation is provided in
Appendix~\ref{app:method_details}.
The forward pass writes only the SMA readout and logsumexp statistics to
HBM. The chunk keys, chunk values, logits, and chunk weights remain
on-chip temporaries and are recomputed in the backward pass.

\paragraph{Training Complexity.}
Let the sequence length be $L$, chunk size be $S$, number of chunks be
$M=\lceil L/S\rceil$, state size be $N$, and head dimension be $P$. In
training, the SMA branch lets each token read from at most
$M-1$ historical chunk state memories. For each
token--chunk pair, the kernel forms a token-conditioned key
$K_{t,c}\in\mathbb{R}^{N}$ and value $V_{t,c}\in\mathbb{R}^{P}$ through
bilinear contractions with $\Delta H_{[c]}\in\mathbb{R}^{N\times P}$.
The resulting compute cost is
\begin{equation}
    O(LMNP)
    =
    O(L^2NP/S).
\end{equation}
Compared with Mamba-2, DART adds this SMA branch on
top of the linear SSD scan, trading additional compute for query-dependent
access to historical compressed states. Compared with token-level attention
implemented by FlashAttention, the SMA branch reads $M$
chunk state memories rather than $L$ token-level KV entries. Its retrieval compute has a
leading-order ratio
\begin{equation}
    \frac{O(L^2NP/S)}{O(L^2P)}
    =
    \frac{N}{S}
\end{equation}
relative to token-level attention per head, and is smaller when $N<S$.

\paragraph{Inference Cache.}
DART is compared against a matched token-level attention baseline during
autoregressive inference. This baseline stores keys and values for each token,
giving a length-dependent KV cache of size $O(2LP)$ per head.
DART instead stores one chunk state memory per chunk, with cache size
$O(MNP)=O(LNP/S)$. The cache ratio to this matched attention baseline is therefore
\begin{equation}
    \frac{LNP/S}{2LP}
    =
    \frac{N}{2S}.
\end{equation}
For example, with chunk size $S=256$ and state size $N=128$, the
chunk state-memory cache is $25\%$ of the token-level KV cache, saving
$75\%$ of the length-dependent cache memory.

\section{Experiments}

We evaluate DART on synthetic MQAR to test associative recall, pretrain it on
the Pile for 100B tokens, and evaluate on general language-modeling and retrieval tasks. 
We further analyze runtime and memory
behavior of the SMA, with more details provided in
Appendix~\ref{app:experimental_details}.

\subsection{Synthetic Tasks: Associative Recall}

Following Mamba-2~\citep{dao2024transformers}, we use synthetic associative
recall to test whether DART improves the ability to look up information
from context. We adopt multi-query associative recall (MQAR), which requires an
autoregressive model to memorize multiple in-context key--value associations
and predict the associated value when queried by a previously seen key
~\citep{arora2024zoology}. We compare DART against Mamba-2 under the
same sequence length, model width, and SSD state size, and include the
Transformer++ attention baseline~\citep{dao2024transformers} under the same
sequence length and model width. The chunk size is set to $S=16$.
For DART, this relatively small chunk size tests whether SMA can retrieve
the stored associations from chunk state memories.

\begin{figure}[htp]
    \centering
    \includegraphics[width=0.9\linewidth]{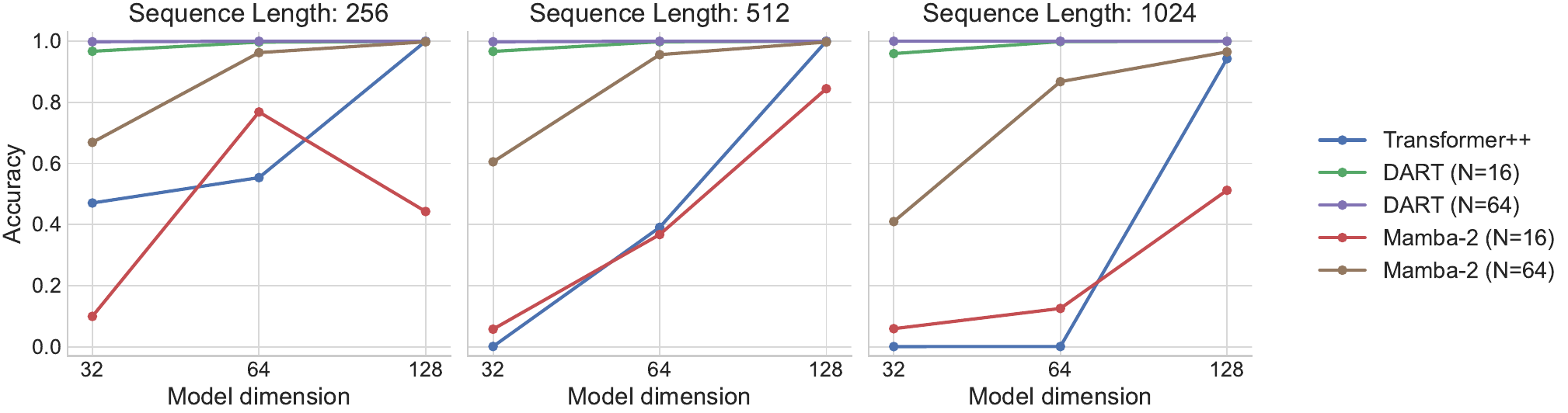}
    \caption{MQAR recall accuracy. Each panel fixes the sequence length $L$ and
    sweeps the model width $d_{\mathrm{model}}$. DART and Mamba-2 are
    evaluated with SSD state sizes $N\in\{16,64\}$. 
    All markers report test accuracy.}
    \label{fig:mqar_accuracy}
\end{figure}

Results are shown in Figure~\ref{fig:mqar_accuracy}. We sweep sequence
length $L\in\{256,512,1024\}$ and model width
$d_{\mathrm{model}}\in\{32,64,128\}$. For Mamba-2 and DART, we evaluate
SSD state sizes $N\in\{16,64\}$. More details are provided in
Appendix~\ref{app:mqar_details}.
DART consistently improves MQAR recall over Mamba-2 under matched model
width and state size. With $N=64$, DART reaches near-perfect accuracy across
all tested sequence lengths and model widths. With the smaller state size
$N=16$, DART also remains highly accurate, and even exceeds Mamba-2 with
$N=64$ in the matched-width configurations. In contrast, Mamba-2 degrades
substantially when either the state size or model width is small, especially at
longer sequence lengths. The Transformer++ baseline is strong when the width is
large enough, but is unreliable in low-width and long-sequence settings. These
results indicate that SMA makes compressed SSD memories much
more accessible than the native Mamba-2 readout, especially when recurrent
state capacity is limited.

\subsection{Language Modeling}

Following standard LLM protocols, we train and evaluate DART on
autoregressive language modeling tasks. We compare pretraining perplexity on
the Pile, zero-shot evaluations on standard downstream tasks, and retrieval
performance on real-world extraction tasks and synthetic NIAH benchmarks.

\paragraph{General Sequence Modeling.}

We start from the baseline Mamba-2 model sizes of 130M, 370M, and
780M parameters, and construct DART by adding the SMA branch to the same backbone shapes. All models are pretrained on the
Pile for 100B tokens, and the resulting models are evaluated with Pile
validation metrics and zero-shot downstream evaluations. Details of the Pile
pretraining setup are provided in Appendix~\ref{app:pile_pretraining_details}.

Table~\ref{tab:pile_pretraining_ppl} reports Pile validation perplexity and
accuracy for the trained models, while Table~\ref{tab:short_downstream}
reports zero-shot downstream results.
For a public full-attention reference trained under a comparable token budget,
we also include Pythia checkpoints~\citep{biderman2023pythia} released at
approximately 100B pretraining tokens in the downstream evaluation.
DART adds fewer than $3\%$ parameters through the SMA branch at these model
sizes. This comparison tests whether the added state-memory retrieval improves
recall and retrieval while preserving general sequence-modeling quality. Across
the evaluated model sizes, DART preserves Pile perplexity, improves Pile
validation accuracy, and obtains higher average zero-shot downstream accuracy
than Mamba-2. The gains are especially clear at the smaller 130M and 370M
scales, indicating that DART preserves general language-modeling capacity.
Most standard downstream examples in Table~\ref{tab:short_downstream} are
short-context inputs, with prompt-answer lengths below 256 tokens. Thus, the
$S=256$ setting mainly evaluates whether training with SMA preserves the
general capability of the SSM branch. The smaller evaluation chunk sizes
$S=64$ and $S=32$ invoke state-memory retrieval on more examples and show that
DART maintains downstream performance when the SMA branch is active at
evaluation time.

\begin{table}[htp]
    \centering
    \caption{Model parameters and Pile validation metrics after 100B-token
    pretraining. DART adds the SMA branch on top of the Mamba-2 backbone,
    resulting in a small parameter increase and competitive Pile validation
    metrics across the completed model scales. Lower perplexity and higher
    accuracy are better.}
    \label{tab:pile_pretraining_ppl}
    \begin{tabular}{lcccccc}
        \hline
        & Mamba-2 & DART & Mamba-2 & DART &
        Mamba-2 & DART \\
        Model scale & 130M & 133M & 370M & 380M &
        780M & 795M \\
        \hline
        Parameters & 128.99M & 132.70M & 368.35M & 378.22M &
        780.16M & 794.95M \\
        Pile PPL $\downarrow$ & 12.36 & 12.25 & 10.18 & 10.10 & 9.19 & 9.20 \\
        Pile Acc. $\uparrow$ & 50.58 & 50.75 & 53.27 & 53.45 & 54.86 & 54.88 \\
        \hline
    \end{tabular}
\end{table}

\begin{table}[htp]
    \centering
    \caption{Zero-shot downstream evaluation of models pretrained on the Pile.
    DART uses $S=256$ as the pretraining and main inference setting. We also
    include $S=64$ and $S=32$ as evaluation-time chunk-size ablations of the
    same trained checkpoint to show how chunk size affects performance.
    LAMBADA PPL is reported as perplexity; all other
    entries are accuracies in percent, and Avg. is the mean over the seven
    accuracy metrics. Pythia models are public full-attention references
    released at approximately 100B pretraining tokens. 
    Best and second-best
    entries within each model-size group are shown in bold and underline,
    respectively.}
    \label{tab:short_downstream}
    \resizebox{\linewidth}{!}{
    \begin{tabular}{lcccccccccc}
        \hline
        Model & Chunk Size & LAMBADA PPL $\downarrow$ & LAMBADA & HellaSwag & PIQA &
        ARC-E & ARC-C & WinoGrande & OpenBookQA & Avg. \\
        \hline
        Mamba-2 130M & -- & \textbf{22.44} & \underline{37.94} &
        32.34 & 62.40 & 39.73 & 23.21 & 49.49 & 28.40 & 39.07 \\
        DART 133M & 256 & \underline{22.77} &
        \textbf{38.35} & \textbf{32.69} & \textbf{62.95} &
        \textbf{41.29} & \textbf{24.23} & \textbf{51.14} &
        \underline{29.80} & \textbf{40.06} \\
        DART 133M & 64 & 25.28 & 36.66 & 32.66 &
        \underline{62.79} & \underline{41.25} & \textbf{24.23} &
        \textbf{51.14} & \underline{29.80} & \underline{39.79} \\
        DART 133M & 32 & 24.13 & 37.07 &
        \underline{32.68} & \underline{62.79} & 41.16 &
        \underline{23.81} & \underline{50.59} & \textbf{30.00} & 39.73 \\
        Pythia 160M & -- & 35.67 & 33.30 & 31.10 & 62.40 &
        40.70 & 23.38 & 50.20 & 26.40 & 38.21 \\
        \hline
        Mamba-2 370M & -- & 13.53 & 46.85 & 38.93 & 66.49 &
        \textbf{45.41} & 24.15 & 51.07 & 29.40 &
        43.18 \\
        DART 380M & 256 & \textbf{12.68} & \textbf{47.45} &
        \textbf{39.17} & \underline{68.72} & 43.43 &
        \textbf{24.66} & 50.99 & \textbf{30.80} &
        \underline{43.60} \\
        DART 380M & 64 & 13.18 & 46.40 & 39.07 &
        \underline{68.72} & 43.43 & \underline{24.57} &
        \textbf{52.17} & \textbf{30.80} & 43.59 \\
        DART 380M & 32 & \underline{12.92} &
        \underline{47.22} & \underline{39.08} & \textbf{68.88} &
        43.52 & 24.40 & \underline{52.09} & \textbf{30.80} &
        \textbf{43.71} \\
        Pythia 410M & -- & 16.20 & 43.76 & 37.09 & 65.56 &
        \underline{44.07} & 24.15 & 51.07 & \underline{30.40} & 42.30 \\
        \hline
        Mamba-2 780M & -- & \underline{9.50} & \textbf{52.32} &
        \textbf{44.42} & 68.34 & 48.11 & 25.94 & 52.88 &
        \textbf{31.40} & 46.20 \\
        DART 795M & 256 & \textbf{9.44} & \underline{52.22} &
        44.03 & 69.10 & \textbf{50.21} & \textbf{27.13} &
        \textbf{54.46} & 30.80 & \textbf{46.85} \\
        DART 795M & 64 & 9.87 & 51.27 & 44.16 &
        \textbf{69.26} & \underline{50.17} & \underline{27.05} &
        \underline{54.22} & \underline{31.20} & 46.76 \\
        DART 795M & 32 & 9.76 & 51.50 & \underline{44.18} &
        \underline{69.15} & \underline{50.17} & \underline{27.05} &
        \textbf{54.46} & 31.00 & \underline{46.79} \\
        Pythia 1B & -- & 10.77 & 50.81 & 42.47 & 67.46 & 47.52 &
        24.91 & 51.78 & 30.80 & 45.11 \\
        \hline
    \end{tabular}
    }
\end{table}


\paragraph{Retrieval Capacity.}
We evaluate retrieval on synthetic needle-in-a-haystack (NIAH) tasks and
real-world tasks covering semi-structured extraction, passage completion, and
question answering; details are provided in
Appendix~\ref{app:downstream_tasks}. As shown in
Table~\ref{tab:retrieval_capacity}, DART gives the clearest gains on
context-conditioned extraction tasks. Across all three model scales, DART
substantially improves SWDE and FDA over Mamba-2. 
The gains on QA-style retrieval tasks are
smaller but generally positive.
DART also improves many NIAH settings that require retrieving information from
the provided context. The largest gains appear on NIAH-Single-2 and
NIAH-Single-3 at 1024 and 2048 tokens, where Mamba-2 often has very low recall
but DART recovers a large fraction of the targets. The hardest 4096-token
NIAH-Single-2/3 settings remain challenging for both models.
Table~\ref{tab:retrieval_capacity} reports DART at its pretraining chunk size
$S=256$ as the main configuration, and includes evaluation-time chunk-size
ablations at $S=64$ and $S=32$ using the same trained checkpoints. At $S=256$,
DART already substantially improves over Mamba-2 across extraction tasks and
many NIAH settings. The chunk-size ablations further measure sensitivity to inference
granularity: smaller chunks improve several extraction-heavy metrics such as
SWDE and FDA, whereas $S=64$ is strongest on some NIAH-Single-2/3 settings.
Pythia remains a strong full-attention reference on several QA and NIAH
columns, indicating that DART narrows the retrieval gap to attention while
operating under the compact-memory constraints of recurrent states.

\begin{table}[htp]
    \centering
    \caption{Retrieval evaluation of models pretrained on the Pile. The
    real-world suite includes extraction-heavy tasks (SWDE, SQuAD Completion,
    and FDA) and QA-style retrieval tasks (TriviaQA, NQ, and DROP). 
    DART uses $S=256$ as the pretraining and main inference setting. We also
    include $S=64$ and $S=32$ as evaluation-time chunk-size ablations of the
    same trained checkpoint to show how chunk size affects performance.
    NIAH reports
    synthetic retrieval accuracy with 500 examples per context length. All
    real-world tasks are evaluated with contains accuracy. 
    Pythia models are public full-attention references released at
    approximately 100B pretraining tokens and are
    evaluated up to their native 2048-token context length. All entries are
    percentages, and higher is better. Best and second-best entries within each
    model-size group are shown in bold and underline, respectively.}
    \label{tab:retrieval_capacity}
    \resizebox{\linewidth}{!}{
    \begin{tabular}{lcccccccccc ccc ccc}
        \hline
        & & \multicolumn{6}{c}{Real-world retrieval} &
        \multicolumn{3}{c}{NIAH-Single-1} &
        \multicolumn{3}{c}{NIAH-Single-2} &
        \multicolumn{3}{c}{NIAH-Single-3} \\
        Model & Chunk Size & SWDE & SQD & FDA & TQA & NQ & DROP &
        1024 & 2048 & 4096 & 1024 & 2048 & 4096 &
        1024 & 2048 & 4096 \\
        \hline
        Mamba-2 130M & -- & 17.1 & 25.6 & 15.4 & \underline{38.4} &
        5.7 & 16.6 & \underline{98.6} & 97.0 &
        \textbf{74.2} & 2.4 & 1.0 & \textbf{1.6} & 1.0 & 1.0 & 1.2 \\
        DART 133M & 256 & 34.9 & 23.3 & 19.7 &
        37.8 & \textbf{9.1} & 17.3 &
        \textbf{100.0} & \textbf{99.4} & 57.2 &
        59.0 & \underline{47.2} & \textbf{1.6} &
        36.0 & \underline{46.6} & \textbf{5.0} \\
        DART 133M & 64 & 44.8 & \underline{26.6} &
        26.2 & \textbf{38.8} & 7.1 & \underline{18.1} &
        \textbf{100.0} & \textbf{99.4} & 65.2 &
        \underline{65.6} & 42.2 & \underline{1.4} &
        \textbf{71.4} & \textbf{47.2} & \underline{4.4} \\
        DART 133M & 32 & \underline{46.1} & \textbf{26.7} &
        \underline{29.0} & \underline{38.4} & 6.4 & \textbf{18.2} &
        \textbf{100.0} & 98.4 & \underline{66.8} &
        50.6 & 27.6 & \underline{1.4} &
        37.0 & 24.8 & 3.0 \\
        Pythia 160M & -- & \textbf{50.9} & 26.5 & \textbf{37.0} &
        35.4 & \underline{9.0} & \textbf{18.2} &
        98.2 & \underline{99.2} & -- & \textbf{98.0} &
        \textbf{83.2} & -- & \underline{52.8} & 25.4 & -- \\
        \hline
        Mamba-2 370M & -- & 33.2 & 29.9 &
        16.6 & 45.9 & 9.5 & \textbf{20.9} &
        \textbf{100.0} & \textbf{100.0} & \underline{95.4} &
        1.4 & 1.4 & \underline{1.8} & 2.6 & 3.4 & 2.8 \\
        DART 380M & 256 & 46.9 & 29.8 &
        31.3 & \underline{46.3} & \underline{12.6} & 18.9 &
        \textbf{100.0} & \textbf{100.0} & \textbf{95.8} &
        13.0 & 14.0 & \textbf{3.2} & 2.8 & 4.4 & \textbf{5.8} \\
        DART 380M & 64 & \underline{56.7} & 31.2 & 45.8 &
        45.3 & 11.6 & 18.9 &
        \underline{99.8} & \underline{99.4} & 86.6 &
        24.0 & 32.6 & \underline{1.8} & 22.8 & 18.4 &
        \underline{4.6} \\
        DART 380M & 32 & 56.2 & \underline{31.5} &
        \underline{50.4} & 44.4 & 10.4 & 19.0 &
        99.6 & 97.0 & 65.8 &
        \underline{36.0} & \underline{35.4} & 1.4 &
        \underline{35.4} & \underline{34.8} & 3.4 \\
        Pythia 410M & -- & \textbf{68.8} & \textbf{32.9} &
        \textbf{62.7} & \textbf{49.1} & \textbf{14.3} & \underline{20.8} &
        99.2 & 99.2 & -- & \textbf{99.6} & \textbf{85.4} & -- &
        \textbf{98.2} & \textbf{91.8} & -- \\
        \hline
        Mamba-2 780M & -- & 39.9 & 32.6 &
        17.2 & 50.5 & 11.8 & 21.3 & 85.8 & 91.2 &
        \underline{99.8} & 40.6 & 24.6 & \textbf{1.8} & 70.6 &
        19.4 & \textbf{3.6} \\
        DART 795M & 256 & 56.5 & 32.4 & 28.9 &
        \underline{52.0} & \underline{16.1} & \underline{23.5} &
        \textbf{100.0} & \textbf{100.0} & \underline{99.8} &
        60.6 & 70.0 & \underline{1.4} & 67.6 & 43.4 & \underline{2.6} \\
        DART 795M & 64 & 63.7 & \textbf{35.5} &
        41.4 & 51.1 & 15.7 & \textbf{23.6} &
        \textbf{100.0} & \textbf{100.0} & \textbf{100.0} &
        \underline{83.0} & \underline{90.6} & \underline{1.4} &
        \textbf{99.6} & 80.6 & 2.4 \\
        DART 795M & 32 & \underline{65.9} & \underline{34.8} &
        \underline{47.9} & 50.4 & 14.5 & 23.4 &
        \textbf{100.0} & \textbf{100.0} & \textbf{100.0} &
        80.6 & 89.4 & \underline{1.4} & \underline{99.2} &
        \underline{88.8} & 2.4 \\
        Pythia 1B & -- & \textbf{68.7} & \underline{34.8} &
        \textbf{66.6} & \textbf{52.5} & \textbf{17.6} & 23.2 &
        \underline{99.0} & \underline{97.4} & -- & \textbf{99.2} &
        \textbf{99.8} & -- & 91.0 & \textbf{97.0} & -- \\
        \hline
    \end{tabular}
    }
\end{table}

\subsection{Ablations and Performance Analysis}

\paragraph{SMA Branch.}
We assess the contribution of the SMA branch through an evaluation-time
ablation. For each trained DART checkpoint, we remove the SMA residual readout
during evaluation and keep only the native SSM branch. The resulting
performance change measures how much associative-recall and retrieval capacity
is provided by the SMA branch beyond the recurrent state
readout. 
Table~\ref{tab:sma_branch_retrieval_ablation} reports the retrieval-task
ablation,
and Table~\ref{tab:sma_branch_mqar_ablation} reports the MQAR ablation.
Removing SMA substantially weakens MQAR and real-world extraction performance,
indicating that the retrieval gains mainly come from making chunk state
memories explicitly addressable. The NIAH results are more task-dependent,
showing that SMA improves access to chunk state memories but does not by itself
solve all synthetic long-context retrieval variants.

\begin{table}[htp]
    \centering
    \caption{Evaluation-time ablation of the SMA branch on NIAH and real-world
    extraction tasks for Pile-pretrained checkpoints. The ablated rows use the
    same trained checkpoints as the full model, but remove the SMA residual
    readout during evaluation. All evaluations use $S=256$, and all entries
    are accuracies in percent.}
    \label{tab:sma_branch_retrieval_ablation}
    \scriptsize
    \resizebox{\linewidth}{!}{
    \begin{tabular}{lccccccccc cc}
        \hline
        & \multicolumn{3}{c}{NIAH-Single-1} &
        \multicolumn{3}{c}{NIAH-Single-2} &
        \multicolumn{3}{c}{NIAH-Single-3} &
        \multicolumn{2}{c}{Real} \\
        Model & 1024 & 2048 & 4096 & 1024 & 2048 & 4096 &
        1024 & 2048 & 4096 & SWDE & FDA \\
        \hline
        DART 133M w/o SMA & 12.8 & 0.0 & 0.0 &
        0.0 & 0.0 & 0.0 & 0.0 & 0.0 & 0.0 &
        4.2 & 0.7 \\
        DART 133M & 100.0 & 99.4 & 57.2 & 59.0 &
        47.2 & 1.6 & 36.0 & 46.6 & 5.0 & 34.9 & 19.7 \\
        DART 380M w/o SMA & 100.0 & 95.8 & 66.2 &
        4.6 & 4.2 & 3.0 & 2.4 & 4.6 & 3.2 &
        13.8 & 7.1 \\
        DART 380M & 100.0 & 100.0 & 95.8 & 13.0 &
        14.0 & 3.2 & 2.8 & 4.4 & 5.8 & 46.9 & 31.3 \\
        DART 795M w/o SMA & 99.8 & 98.0 & 90.8 &
        11.6 & 13.2 & 3.0 & 29.6 & 18.6 & 3.2 &
        17.3 & 3.0 \\
        DART 795M & 100.0 & 100.0 & 99.8 & 60.6 &
        70.0 & 1.4 & 67.6 & 43.4 & 2.6 & 56.5 & 28.9 \\
        \hline
    \end{tabular}
    }
\end{table}

\paragraph{Sharing the SSM Read Vector.}
DART uses the native Mamba-2 read vector $C_t$ as the value-side read vector in
the SMA branch. This design aligns the SMA readout with the SSM branch
when decoding historical chunk state memories. We train otherwise matched
variants that replace $C_t$ with an independently parameterized SMA read vector
$C_t^{\mathrm{SMA}}=U_tW_C^{\mathrm{SMA}}$.
Table~\ref{tab:c_sharing_ablation} reports the MQAR
comparison in the memory-constrained setting $N=16$.
The shared design consistently succeeds across widths and sequence lengths,
whereas the independently parameterized read vector fails to learn useful
recall behavior in this setting and stays at zero accuracy.
This shows that aligning the SMA value readout with the native SSM readout is
important for decoding historical chunk state memories.

\begin{table}[htp]
    \centering
    \begin{minipage}[t]{0.44\linewidth}
        \centering
        \caption{Evaluation-time ablation of the SMA branch on MQAR. Each row
        uses a DART model with SSD state size $N=16$ and chunk size $S=16$.
        All entries are test accuracies in percent.}
        \label{tab:sma_branch_mqar_ablation}
        \scriptsize
        \resizebox{\linewidth}{!}{
        \begin{tabular}{lcccc}
            \hline
            Model & $d_{\mathrm{model}}$ & $L=256$ & $L=512$ & $L=1024$ \\
            \hline
            DART w/o SMA & 32 & 12.40 & 0.30 & 0.11 \\
            DART & 32 & 96.70 & 96.67 & 95.94 \\
            DART w/o SMA & 64 & 0.55 & 1.07 & 0.06 \\
            DART & 64 & 99.70 & 99.83 & 99.88 \\
            DART w/o SMA & 128 & 0.23 & 0.08 & 0.08 \\
            DART & 128 & 99.94 & 99.99 & 99.98 \\
            \hline
        \end{tabular}
        }
    \end{minipage}
    \hfill
    \begin{minipage}[t]{0.52\linewidth}
        \centering
        \caption{Ablation of $C_t$ sharing on MQAR with SSD state size
        $N=16$ and chunk size $S=16$. All entries are test accuracies in
        percent.}
        \label{tab:c_sharing_ablation}
        \scriptsize
        \resizebox{\linewidth}{!}{
        \begin{tabular}{lcccc}
            \hline
            Model & $d_{\mathrm{model}}$ & $L=256$ & $L=512$ & $L=1024$ \\
            \hline
            DART, shared $C_t$ & 32 & 96.70 & 96.67 & 95.94 \\
            DART, independent $C_t^{\mathrm{SMA}}$ & 32 & 0.00 & 0.00 & 0.00 \\
            DART, shared $C_t$ & 64 & 99.70 & 99.83 & 99.88 \\
            DART, independent $C_t^{\mathrm{SMA}}$ & 64 & 0.00 & 0.00 & 0.00 \\
            DART, shared $C_t$ & 128 & 99.94 & 99.99 & 99.98 \\
            DART, independent $C_t^{\mathrm{SMA}}$ & 128 & 0.00 & 0.00 & 0.00 \\
            \hline
        \end{tabular}
        }
    \end{minipage}
\end{table}

\paragraph{Kernel Forward Time and Inference Cache.}
We evaluate the system efficiency of the SMA branch in terms of kernel forward
time and inference cache size. For runtime, we report forward time for the
kernels under matched batch size, head dimension, state size,
chunk size, and bf16 inputs on a single A800 80GB GPU. The comparison measures
FlashAttention-2 with precomputed $Q,K,V$, SSD with precomputed
$X,\Delta,A,B,C$, and SMA with precomputed $Q,E,C$, chunk state memories
$\{\Delta H_{[c]}\}$.
For inference, we compare the
persistent decode cache associated with these sequence modules: attention KV
cache for FlashAttention-2, conv/recurrent state cache for SSD, and
chunk state-memory cache for SMA. As discussed in
Section~\ref{sec:efficient_kernels}, the ratio of the chunk state-memory cache
to the matched attention KV cache is $N/(2S)$ per head, giving a $25\%$ ratio when
$N=128$ and $S=256$.

\begin{figure}[htbp]
    \centering
    \includegraphics[width=\linewidth]{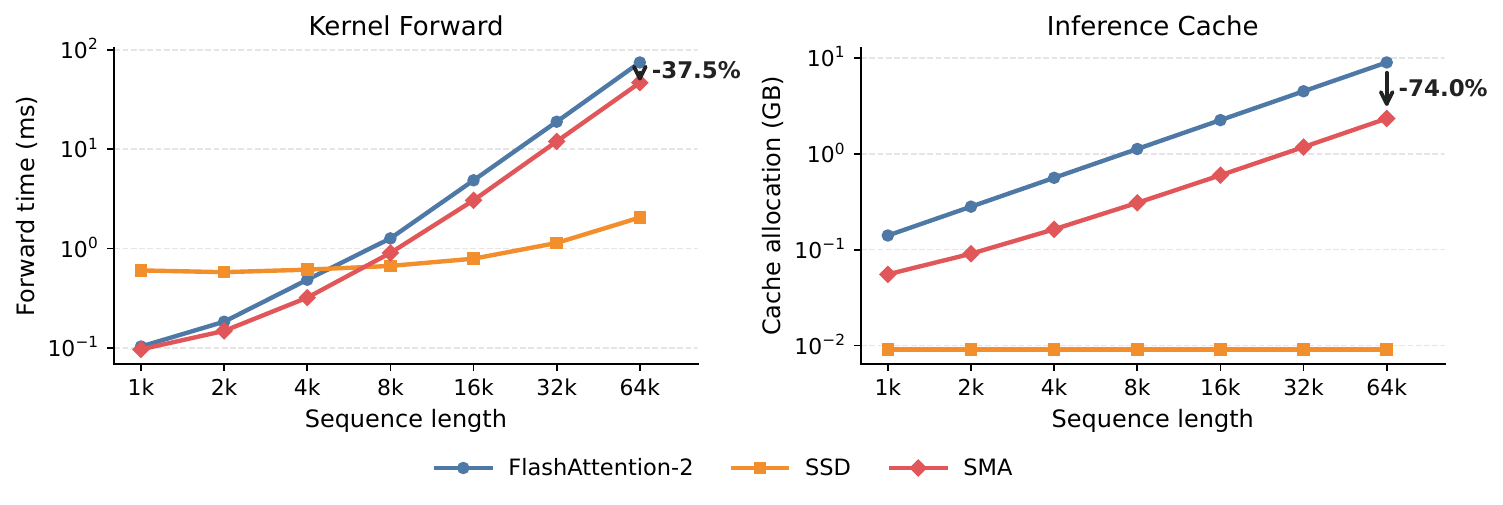}
    \caption{Kernel forward time and inference cache comparison.
    \textbf{Left}: forward time of the kernels on an A800
    80GB GPU with batch size 1, 24 heads, head dimension 64, SSD state size
    128, chunk size 256, and bf16 inputs. \textbf{Right}: measured CUDA
    allocation for persistent inference cache tensors in a 24-layer, 24-head
    model. The results show that SMA consistently has
    lower kernel forward time than FlashAttention-2 in this setting, while also
    requiring substantially smaller persistent inference cache than token-level
    attention.}
    \label{fig:delay_mamba_efficiency}
\end{figure}



\section{Conclusion}

We presented DART, an architecture that augments Mamba-2 with state-memory
attention over compact recurrent states. The Mamba-2 chunked scan produces
chunk state contributions that DART retains as chunk state memories. SMA
decodes token-conditioned keys and values from these memories and adds its
readout as a gated correction to the SSM branch.
Experiments on MQAR,
Pile pretraining, downstream evaluation, retrieval benchmarks, and efficiency
analysis show that DART improves associative recall and retrieval while
preserving general language-modeling quality.
These results demonstrate that recurrent compression and attention-style access 
can be supported by the same recurrent-state representation.



\bibliography{iclr2026_conference}
\bibliographystyle{iclr2026_conference}

\appendix

\section{Mamba-2 Chunked Scan}
\label{app:mamba2}

This appendix reviews the chunked SSD scan used by Mamba-2. Let
$I_c=\{a_c,\ldots,b_c\}$ denote the $c$-th chunk, and let
$H_{[c]}=H_{b_c}$ be its chunk-boundary state. The
token-level recurrence can be written as
\begin{equation}
    H_t = A_t H_{t-1} + B_t^\top X_t,
    \qquad
    Y_t = C_t H_t .
\end{equation}
For compactness, define
\begin{equation}
    A_{u:v}
    =
    A_v A_{v-1}\cdots A_u
    =
    \prod_{k=u}^{v} A_k,
    \qquad u\leq v,
\end{equation}
and set $A_{u:v}=I$ when $u>v$. 
Thus a value written at position $s$ reaches position $t$ through
$A_{s+1:t}$.

\paragraph{Chunked Scan.}
Using arrow notation to indicate that the decay has been absorbed toward the
chunk boundary,
\begin{equation}
    \overrightarrow{H}_{[c-1]}
    =
    A_{a_c:b_c}H_{[c-1]},
    \qquad
    \overleftarrow{C}_t
    =
    C_t A_{a_c:t},
\end{equation}
the Mamba-2 chunked scan computes the chunk-boundary state and decomposes each token
output into intra- and inter-chunk terms:
\begin{equation}
    H_{[c]}
    =
    \overrightarrow{H}_{[c-1]}
    +
    \Delta H_{[c]},
    \label{eq:mamba2_chunk_state}
\end{equation}
\begin{equation}
    Y_t
    =
    Y_t^{\mathrm{intra}}
    +
    Y_t^{\mathrm{inter}},
    \qquad t\in I_c .
\end{equation}
Here $\Delta H_{[c]}$ is the chunk state contribution to the chunk-boundary
state:
\begin{equation}
    \Delta H_{[c]}
    =
    \sum_{s\in I_c}
    A_{s+1:b_c} B_s^\top X_s .
    \label{eq:mamba2_delta_h_chunk}
\end{equation}
The zero-initial-state output is
\begin{equation}
    Y_t^{\mathrm{intra}}
    =
    \sum_{s\in I_c,\,s\leq t}
    C_t A_{s+1:t} B_s^\top X_s ,
    \qquad t\in I_c .
\end{equation}
The inter-chunk readout is
\begin{equation}
    Y_t^{\mathrm{inter}}
    =
    \overleftarrow{C}_t H_{[c-1]} .
\end{equation}
Thus the sequence computation is reduced to chunk-local structured attention
blocks, followed by a scan over the compact chunk-boundary states
$\{H_{[c]}\}$. The matrices $\Delta H_{[c]}$ are the chunk state contributions
retained by DART as chunk state memories.

\paragraph{Matrix Form.}
Stacking the tokens in a chunk gives the same decomposition in block form:
\begin{equation}
    Y_{[c]}
    =
    \overleftarrow{C}_{[c]}H_{[c-1]}
    +
    \mathcal{L}_{[c]} X_{[c]},
\end{equation}
where the lower-triangular SSD block is
\begin{equation}
    \mathcal{L}_{[c]}[t,s]
    =
    \begin{cases}
        C_t A_{s+1:t}B_s^\top, & s\leq t,\quad s,t\in I_c,\\
        0, & s>t,\quad s,t\in I_c .
    \end{cases}
\end{equation}

\section{Method Details}
\label{app:method_details}

\subsection{Normalization in DART}

In the implemented architecture, the state-memory attention uses a normalized
score branch:
\begin{equation}
    \bar Q_t
    =
    \operatorname{RMSNorm}(Q_t),
    \qquad
    \bar E_t
    =
    \operatorname{RMSNorm}(E_t).
\end{equation}
For each chunk state memory, define
\begin{equation}
    r_{c,n}
    =
    \left(
    \frac{1}{P}
    \sum_{p=1}^{P}
    (\Delta H_{[c]})_{n,p}^2
    +\epsilon
    \right)^{-1/2},
    \qquad
    K_{t,c}
    =
    r_c\odot(\Delta H_{[c]}\bar E_t),
\end{equation}
where $r_c\in\mathbb{R}^{N\times 1}$ is the inverse RMS of
$\Delta H_{[c]}$ along the $P$ axis. This stabilizes the score branch
without materializing a normalized copy of $\Delta H_{[c]}$. The
normalized chunk logit and SMA readout are
\begin{equation}
    \bar\ell_{t,c}
    =
    \frac{\bar Q_tK_{t,c}}{\sqrt N},
    \qquad
    g_{t,c}
    =
    \operatorname{softmax}_{c<c(t)}(\bar\ell_{t,c}),
    \qquad
    \bar R_t
    =
    \sum_{c<c(t)}g_{t,c}V_{t,c}.
\end{equation}

\subsection{Multi-Value Details}

Let $\mathcal{B}$ denote the batch size, $L$ the sequence length, $H$ the
number of heads, $P$ the value dimension, $N$ the SSD state size,
$M=\lceil L/S\rceil$ the number of chunks, and $G$ the number of grouped
projections. In the current experiments, DART uses grouped
$B$, $C$, $Q$, and $E$ with $G=1$.

\begin{table}[htbp]
    \centering
    \caption{Compact tensor shapes used by DART with MVA-style grouped
    projections.}
    \label{tab:mva_tensor_shapes}
    \begin{tabular}{lc}
        \hline
        Tensor & Shape \\
        \hline
        $X$ & $\mathcal{B} \times L \times H \times P$ \\
        $\Delta H$ & $\mathcal{B} \times M \times H \times N \times P$ \\
        SSM $B$, $C$, $Q$ & $\mathcal{B} \times L \times G \times N$ \\
        $E$ & $\mathcal{B} \times L \times G \times P$ \\
        \hline
    \end{tabular}
\end{table}

\section{FlashAttention Background}
\label{app:flash_attention}

FlashAttention reduces the memory cost of exact attention by
matching the attention computation to the GPU memory
hierarchy~\citep{dao2022flashattention,dao2024flashattention}. Instead of
materializing the full score matrix in HBM, it streams tiles of keys and
values through on-chip SRAM. Here a tile denotes a small contiguous block of
query, key, or value vectors processed as a matrix block. For a fixed query
tile $Q_i$, the kernel iterates over key-value tiles $(K_j,V_j)$, computes the local scores
$S_{ij}=Q_iK_j^\top$, updates the softmax statistics online, and accumulates
the output tile.

\paragraph{Online Softmax.}
The key idea is to compute the stable softmax while reading a vector in
pieces. For a score vector $x$, let $m=\max_k x_k$. The stable softmax is
\begin{equation}
    \operatorname{softmax}(x)_j
    =
    \frac{e^{x_j-m}}{\sum_k e^{x_k-m}} .
\end{equation}
Suppose the entries of $x$ are processed one by one. After seeing
$x_1,\ldots,x_{j-1}$, online softmax maintains
\begin{equation}
    m^{(j-1)}
    =
    \max_{k< j} x_k,
    \qquad
    \ell^{(j-1)}
    =
    \sum_{k< j} e^{x_k-m^{(j-1)}},
\end{equation}
with $m^{(0)}=-\infty$ and $\ell^{(0)}=0$. Here $m^{(j-1)}$ records the
largest score seen so far, and $\ell^{(j-1)}$ records the softmax denominator
under the current maximum. After seeing a new
entry $x_j$, the statistics are updated as
\begin{equation}
    m^{(j)}
    =
    \max\!\left(m^{(j-1)},x_j\right),
    \qquad
    \ell^{(j)}
    =
    e^{m^{(j-1)}-m^{(j)}}\ell^{(j-1)}
    +
    e^{x_j-m^{(j)}}.
\end{equation}
When softmax is used to form a weighted value sum, the numerator accumulator
is rescaled by the same factor and then receives the new value contribution:
\begin{equation}
    O^{(j)}
    =
    e^{m^{(j-1)}-m^{(j)}}O^{(j-1)}
    +
    e^{x_j-m^{(j)}}v_j .
\end{equation}
After all entries have been processed, the output is $O/\ell$. FlashAttention
applies this same procedure to every query row while streaming key-value
tiles.

\paragraph{Recomputation.}
FlashAttention uses recomputation to avoid storing attention score and
probability tiles. During the forward pass, the kernel computes each local
score tile $S_{ij}=Q_iK_j^\top$ on chip, updates the online softmax
statistics, accumulates the output, and stores only the output together with
one logsumexp value per query row. During the backward pass, the same local
score tile $S_{ij}=Q_iK_j^\top$ is recomputed from $Q_i$ and $K_j$, and the
stored logsumexp is used to recover the corresponding softmax probabilities.

Concretely, for a score vector $x$, let $m=\max_j x_j$. Its logsumexp can be
computed in the stable form
\begin{equation}
    \operatorname{LSE}(x)
    =
    m+\log \sum_j e^{x_j-m}.
\end{equation}
The softmax probability can then be written as
\begin{equation}
    \operatorname{softmax}(x)_j
    =
    \frac{e^{x_j-m}}{\sum_k e^{x_k-m}}
    =
    e^{x_j-\operatorname{LSE}(x)} .
\end{equation}
Thus recovering the softmax probabilities only requires the recomputed score
vector and its logsumexp. The online softmax statistics provide this quantity
because $\ell=\sum_j e^{x_j-m}$ is the stable softmax denominator, and
\begin{equation}
    L_{\mathrm{lse}}
    =
    m+\log \ell .
\end{equation}
For a recomputed score tile $S_{ij}$, the probability tile is recovered as
\begin{equation}
    P_{ij}
    =
    \exp(S_{ij}-L_{\mathrm{lse},i}).
\end{equation}
This trades additional arithmetic for lower activation memory, because the
attention probabilities and intermediate score tiles do not need to be stored
in HBM.

\section{Experiment Configuration Details}
\label{app:experimental_details}

\subsection{Synthetic Tasks: Associative Recall}
\label{app:mqar_details}

Following the harder MQAR variant used in Mamba-2~\citep{dao2024transformers}, non-special positions are filled with random
vocabulary tokens rather than zeros.  
This prevents the model from relying on a sparse blank background and makes the task closer to a dense language-modeling sequence.

\paragraph{Task construction.}
For a sequence of length $L$, let $R$ be the number of key--value pairs.  Each
example samples $R$ distinct keys
$k_1,\ldots,k_R$ from the key vocabulary and $R$ distinct values
$v_1,\ldots,v_R$ from a disjoint value vocabulary.  The first $2R$ positions
contain the source associations:
\begin{equation}
    x_{2i-1}=k_i,\qquad x_{2i}=v_i,\qquad i=1,\ldots,R .
\end{equation}
The remaining sequence provides alternating candidate query slots at positions
$\{2R+1,2R+3,\ldots,L-1\}$. We choose $R$ distinct slots and insert a randomly
permuted copy of the source keys into them.
At a query position containing key $k_i$, the target is its associated value
$v_i$.
Training and evaluation use causal model forward passes with supervised labels
only at query positions: the model predicts the associated value from the
prefix ending at the query key, while all other labels are ignored.


\paragraph{Lengths and number of pairs.}
We evaluate sequence lengths
\begin{equation}
    L\in\{256,512,1024\},
\end{equation}
and use $R=L/4$ key--value pairs at evaluation time.  Since the number of query
slots is also $L/4$, all query slots are occupied in this setting.

\paragraph{Optimization.}
The training set contains $2^{18}$ examples per run.  Validation and test
examples use the full evaluation setting $R=L/4$, while training uses a
four-stage pair-count curriculum
$R\in\{L/16,L/8,3L/16,L/4\}$ with $8$ epochs per stage.  Thus each run uses
$32$ epochs in total. 
The effective batch size is chosen so that each
optimization step contains $2^{18}\approx 0.25M$ tokens:
$1024$ sequences for $L=256$, $512$ sequences for $L=512$, and $256$ sequences
for $L=1024$.  We train with AdamW, gradient clipping at norm $1.0$, and a
linear learning-rate decay over the full training run.  For each model
configuration, we run peak learning rates $10^{-3}$ and $3\times10^{-4}$ and
report the run with the better validation accuracy.

\paragraph{Model settings.}
All MQAR models use two layers and untied input and output embeddings.  DART
and Mamba-2 use chunk size $16$.  We sweep sequence length, SSD state size, and
model width:
\begin{equation}
    L\in\{256,512,1024\},\qquad
    N\in\{16,64\},\qquad
    d_{\mathrm{model}}\in\{32,64,128\}.
\end{equation}
The Transformer++ attention baseline uses the same $(L,d_{\mathrm{model}})$ grid.

\subsection{Pile Pretraining}
\label{app:pile_pretraining_details}

We pretrain all language models on the Pile for approximately $100B$ tokens
with context length $2048$.
The training batch is fixed to $1024$ sequences per optimizer step. 
We train with AdamW, $\beta=(0.9,0.95)$, weight
decay $0.1$, gradient clipping at norm $1.0$, bf16 mixed precision, and a
cosine learning-rate schedule with $1\%$ warmup and minimum learning-rate ratio
$0.1$.

\begin{table}[htp]
    \centering
    \caption{Pile pretraining settings. DART uses the same Mamba-2 backbone
    shape as the corresponding Mamba-2 baseline and adds the SMA branch.}
    \label{tab:pile_pretraining_config}
    \resizebox{0.92\linewidth}{!}{
    \begin{tabular}{lcccccc}
        \hline
        Model & Parameters & Layers & Hidden size & State size &
        Chunk size & Peak LR \\
        \hline
        Mamba-2 130M & 128.99M & 24 & 768 & 128 & 256 & $6.0\times10^{-4}$ \\
        DART 133M & 132.70M & 24 & 768 & 128 & 256 & $6.0\times10^{-4}$ \\
        Mamba-2 370M & 368.35M & 48 & 1024 & 128 & 256 & $3.0\times10^{-4}$ \\
        DART 380M & 378.22M & 48 & 1024 & 128 & 256 & $3.0\times10^{-4}$ \\
        Mamba-2 780M & 780.16M & 48 & 1536 & 128 & 256 & $2.5\times10^{-4}$ \\
        DART 795M & 794.95M & 48 & 1536 & 128 & 256 & $2.5\times10^{-4}$ \\
        \hline
    \end{tabular}
    }
\end{table}

All Mamba-2 and DART models use head dimension $64$,
one shared $B,C$ group, convolution width $4$, and RMSNorm. DART uses chunk
size $S=256$ during pretraining; different inference chunk sizes are evaluated
from the same trained model in the downstream and retrieval experiments. In
all language-modeling experiments, DART uses grouped $Q$, $B$, $C$, and $E$
projections and a scalar SMA residual gate per token for parameter efficiency.

We include public Pythia checkpoints~\citep{biderman2023pythia} as
full-attention references. To match our 100B-token training budget, we evaluate
the publicly released Pythia checkpoints corresponding to approximately 100B
pretraining tokens. These models follow the GPT-NeoX Transformer architecture
with rotary position embeddings, rotary fraction $0.25$, and native context
length $2048$. 
The official architecture settings are summarized in
Table~\ref{tab:pythia_reference_config}.

\begin{table}[htp]
    \centering
    \caption{Pythia reference model settings from the official configurations.
    All models use full attention with RoPE and native context length $2048$.}
    \label{tab:pythia_reference_config}
    \resizebox{0.8\linewidth}{!}{
    \begin{tabular}{lccccc}
        \hline
        Model & Layers & Hidden size & Attention heads & Head dim. &
        Context length \\
        \hline
        Pythia 160M & 12 & 768 & 12 & 64 & 2048 \\
        Pythia 410M & 24 & 1024 & 16 & 64 & 2048 \\
        Pythia 1B & 16 & 2048 & 8 & 256 & 2048 \\
        \hline
    \end{tabular}
    }
\end{table}

\subsection{Downstream Evaluation Tasks}
\label{app:downstream_tasks}

\paragraph{Evaluation protocol.}
All downstream and retrieval evaluations are zero-shot. We use the standard
multiple-choice or completion formulation for the general language-modeling
tasks. For the retrieval suite, SWDE, SQuAD Completion, and FDA use the
completion-format evaluation from the Based benchmark, while TriviaQA, Natural
Questions, and DROP are evaluated in the cloze-style format. These retrieval tasks are
scored by contains accuracy: an example
is counted as correct if the generated response contains any accepted answer
string. For NIAH, we use the RULER single-needle generators with 500 examples
per context length and report accuracy separately at 1024, 2048, and 4096
tokens. 
All generation-based retrieval tasks are evaluated with greedy decoding.
The
real-world retrieval tasks use maximum context length 2048 and generate up to
48 tokens, while NIAH uses maximum context length 4096 and generates up to 128
tokens. 

\paragraph{General Language Modeling Tasks.}

\begin{itemize}
    \item \textbf{Pile.} The Pile~\citep{gao2020pile} is a large-scale
    English corpus that combines web text, books, code, academic papers, and
    other written sources. We report language-modeling perplexity to measure
    general next-token prediction quality on this broad text distribution.
    \item \textbf{LAMBADA.} LAMBADA~\citep{paperno2016lambada} evaluates long-context word prediction by
    asking the model to infer the final word of a passage, with results reported
    as both perplexity and exact-match accuracy.
    \item \textbf{HellaSwag.} HellaSwag~\citep{zellers2019hellaswag} is a commonsense sentence-completion
    benchmark where the model selects the most plausible continuation among
    adversarially constructed choices.
    \item \textbf{PIQA.} PIQA~\citep{bisk2020piqa} measures physical commonsense reasoning by asking
    the model to choose the more plausible solution for everyday situations.
    \item \textbf{ARC-Easy.} ARC-Easy~\citep{clark2018think} evaluates multiple-choice elementary
    science questions that are usually solvable with relatively direct factual
    knowledge.
    \item \textbf{ARC-Challenge.} ARC-Challenge~\citep{clark2018think} uses the harder subset of
    multiple-choice science questions that more often require multi-step
    reasoning or stronger commonsense knowledge.
    \item \textbf{WinoGrande.} WinoGrande~\citep{sakaguchi2021winogrande} evaluates pronoun and coreference
    resolution through commonsense sentence pairs designed to reduce simple
    lexical shortcuts.
    \item \textbf{OpenBookQA.} OpenBookQA~\citep{mihaylov2018can} tests open-book-style elementary
    science reasoning by combining a short scientific fact with commonsense
    inference over multiple-choice answers.
\end{itemize}

\paragraph{Retrieval Tasks.}
Following the retrieval suite adopted by Mamba-3~\citep{lahoti2026mamba}, we
include real-world retrieval tasks from the evaluation suite of Based and Just
Read Twice~\citep{arora2024simple,arora2024just}, together with synthetic
needle-in-a-haystack tasks from RULER~\citep{hsieh2024ruler}.
\begin{itemize}
    \item \textbf{SWDE.} SWDE~\citep{hao2011one} evaluates semi-structured web information
    extraction by asking the model to recover an attribute value from an HTML
    page or page-derived text.
    \item \textbf{SQuAD Completion.} SQuAD
    Completion~\citep{rajpurkar2018know,arora2024simple} converts extractive
    reading comprehension into a cloze-style completion task where the model
    generates the answer span from the provided passage and query context.
    \item \textbf{FDA.} FDA~\citep{arora2024simple} evaluates information extraction from regulatory
    documents by asking the model to generate the value associated with a
    requested key in a document chunk.
    \item \textbf{TriviaQA.} TriviaQA~\citep{joshi2017triviaqa} measures open-domain factual retrieval by
    asking the model to generate a short answer to a trivia question.
    \item \textbf{Natural Questions.} Natural Questions~\citep{kwiatkowski2019natural} evaluates open-domain
    question answering over naturally occurring search queries with short
    answer targets.
    \item \textbf{DROP.} DROP~\citep{dua2019drop} tests paragraph-level retrieval and discrete
    reasoning by asking the model to answer questions whose evidence is located
    in a passage.
    \item \textbf{NIAH-Single-1.} NIAH-Single-1 is a synthetic single-needle
    retrieval task that places a word-to-number fact in a repeated-text
    haystack and asks the model to recover the number.
    \item \textbf{NIAH-Single-2.} NIAH-Single-2 is a synthetic single-needle
    retrieval task that places a word-to-number fact in an essay-style haystack
    and asks the model to recover the number.
    \item \textbf{NIAH-Single-3.} NIAH-Single-3 is a synthetic single-needle
    retrieval task that places a word-to-UUID fact in an essay-style haystack
    and asks the model to recover the UUID.
\end{itemize}

\end{document}